\documentclass[12pt]{article}

\usepackage{arxiv}

\usepackage[utf8]{inputenc} 
\usepackage[T1]{fontenc}    

\usepackage{hyperref}       
\usepackage{url}            
\usepackage{booktabs}       
\usepackage{amsfonts}       
\usepackage{nicefrac}       
\usepackage{microtype}      
\usepackage{graphicx}

\usepackage{doi}

\usepackage[affil-it]{authblk}  
\usepackage{placeins}
\usepackage{float}

\usepackage[acronym]{glossaries}
\makeglossaries

\usepackage{amsmath}
\counterwithout{table}{section}  
\usepackage{chngcntr}
\counterwithout{figure}{section}

\usepackage{booktabs}       

\usepackage{cite}

\newacronym{sls}{SLS}{Stroboscopic Light Stimulation}
\newacronym{vh}{VH}{Visual Hallucination}
\newacronym{llm}{LLM}{Large Language Model}
\newacronym{nlp}{NLP}{Natural Language Processing}
\newacronym{asc}{ASC}{Altered States of Consciousness}
\newacronym{tm}{TM}{Topic Modelling}
\newacronym{hs}{HS}{High Sensory}
\newacronym{dl}{DL}{Deep Listening}
\newacronym{lda}{LDA}{Latent Dirichlet Allocation}
\newacronym{umap}{UMAP}{Uniform Manifold Approximation and Projection}
\newacronym{hdbscan}{HDBSCAN}{Hierarchical Density-Based Spatial Clustering of Applications with Noise}

\title{Mapping of Subjective Accounts into Interpreted Clusters (MOSAIC): Topic Modelling and LLM applied to stroboscopic phenomenology}
\date{} 					

\author{
\href{https://orcid.org/0009-0006-4548-5349}{\includegraphics[scale=0.06]{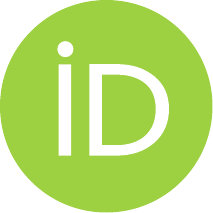}\hspace{1mm} Romy~Beauté}\thanks{\textsuperscript{}Corresponding author: r.beaut@sussex.ac.uk}
\textsuperscript{1,2},
David J. Schwartzman\textsuperscript{1},
Guillaume Dumas\textsuperscript{3,5},
Jennifer Crook\textsuperscript{4},
Fiona Macpherson\textsuperscript{6},
Adam B. Barrett\textsuperscript{1},
Anil K. Seth\textsuperscript{1,5}
}

\affil{\footnotesize\textsuperscript{1}Sussex Center for Consciousness Science, University of Sussex, UK, \textsuperscript{2}be.AI, University of Sussex, \textsuperscript{3}CHUSJ Azrieli Research Center/Mila-Quebec AI Institute, University of Montreal, Canada, \textsuperscript{4}Collective Act, London, UK, \textsuperscript{5}Canadian Institute for Advanced Research, Toronto, Canada, \textsuperscript{6}Centre for the Study of Perceptual Experience, University of Glasgow, UK}

\renewcommand{\shorttitle}{\textit{MOSAIC: Mapping Of Subjective Accounts into Interpreted Clusters} }

\usepackage{setspace}
\hypersetup{
pdftitle={Mapping Subjective Reports via Topic Modelling and LLM Interpretation in Stroboscopically-Induced Phenomena},
pdfsubject={q-bio.NC, q-bio.QM},
pdfauthor={Romy Beauté},
pdfkeywords={Topic Modelling, Computational Phenomenology, Stroboscopic Light Stimulations, Visual Hallucinations, Altered States of Consciousness, Large Language Model, Dreamachine},
}

\begin{document}
\maketitle

\begin{abstract}
Stroboscopic light stimulation (SLS) on closed eyes typically induces simple visual hallucinations (VHs), characterised by vivid, geometric and colourful patterns. A dataset of 862 sentences, extracted from 422 open subjective reports, was recently compiled as part of the Dreamachine programme\footnote{\url{https://dreamachine.world/}}(Collective Act, 2022), an immersive multisensory experience that combines SLS and spatial sound in a collective setting. Although open reports extend the range of reportable phenomenology, their analysis presents significant challenges, particularly in systematically identifying patterns. To address this challenge, we implemented a data-driven approach leveraging Large Language Models and Topic Modelling to uncover and interpret latent experiential topics directly from the Dreamachine's text-based reports. Our analysis confirmed the presence of simple VHs typically documented in scientific studies of SLS, while also revealing experiences of altered states of consciousness and complex hallucinations. Building on these findings, our computational approach expands the systematic study of subjective experience by enabling data-driven analyses of open-ended phenomenological reports, capturing experiences not readily identified through standard questionnaires. By revealing rich and multifaceted aspects of experiences, our study broadens our understanding of stroboscopically-induced phenomena while highlighting the potential of Natural Language Processing and Large Language Models in the emerging field of computational (neuro)phenomenology. More generally, this approach provides a practically applicable methodology for uncovering subtle hidden patterns of subjective experience across diverse research domains.
\end{abstract}


\section*{List of Abbreviations}
\begin{description}
\item[ASC] Altered States of Consciousness 
\item[DL] Deep Listening
\item[HDBSCAN] Hierarchical Density-Based Spatial Clustering of Applications with Noise
\item[HS] High Sensory
\item[LDA] Latent Dirichlet Allocation
\item[LLM] Large Language Model
\item[NLP] Natural Language Processing
\item[SLS] Stroboscopic Light Stimulation
\item[TM] Topic Modelling
\item[UMAP] Uniform Manifold Approximation and Projection
\item[VH] Visual Hallucination
\end{description}

\newpage

\section{INTRODUCTION}

\gls{sls} has long been recognised for its ability to transiently induce a wide range of \glspl{vh} \cite{hewitt_stroboscopically_nodate}. Under closed-eyes conditions \gls{sls} reliably evokes simple (elementary) \glspl{vh}, characterised by colourful geometric patterns that are devoid of semantic content \cite{allefeld_flicker-light_2011,schwartzman_increased_2019,bartossek_altered_2021}. These  \glspl{vh} resemble those observed in other contexts such as psychedelic states \cite{ter_meulen_stroboscope_2009,roseman_lsd_2016,schmidt_altered_2018} and hypnagogic experiences \cite{bartossek_altered_2021}, and pathological conditions such as migraine aura \cite{schott_exploring_2007,cowan_visual_2022}, epileptic seizures \cite{panayiotopoulos_elementary_1994}, Parkinson's Disease \cite{zarkali_flickering_2019,barnes_visual_2001}, Charles Bonnet Syndrome \cite{jan_visual_2012} and Psychosis \cite{sumich_reduction_2018}. 
Although less common, \gls{sls} have been shown to also give rise to complex\glspl{vh} that incorporate figurative or semantic elements, such as realistic scenes, objects, and faces \cite{shenyan_visual_2024,hewitt_stroboscopically_nodate}, potentially reflecting higher-order cortical involvement \cite{howard_anatomy_1998,megevand_seeing_2014}.
Beyond visual phenomena, \gls{sls} can also alter mood, arousal, and the subjective sense of time passing \cite{ter_meulen_stroboscope_2009,bartossek_altered_2021}. The ability of \gls{sls} to elicit diverse subjective experiences under controlled laboratory conditions provides a unique opportunity to investigate the neural and psychological mechanisms underlying hallucinations \cite{pearson_sensory_2016,rogers_hallucinations_2021},
cognitive changes \cite{herrmann_human_2001} and \gls{asc} \cite{allefeld_flicker-light_2011,rule_model_2011,schwartzman_increased_2019,smythies_stroboscopic_1959}. A crucial first step towards this goal is to comprehensively characterise the subjective phenomena \gls{sls} produces. 

\glspl{asc} generally encompass changes in both overall conscious state and specific experiential contents, including perception, affect, cognition, and self-experience  \cite{bartossek_altered_2021,schwartzman_increased_2019}. To our knowledge, only one study has investigated the potential of SLS to induce \gls{asc}, but did not report a significant alteration in consciousness   \cite{bartossek_altered_2021}.  Specifically, Bartossek et al., (2021) examined the effects of 3~Hz and 10~Hz periodic \gls{sls} using well-validated questionnaires, including the \gls{asc} Rating Scale (5D-\gls{asc}/11D-\gls{asc}) \cite{dittrich_standardized_1998,dittrich_5d-asc_2010,studerus_psychometric_2010} and the Phenomenology of Consciousness Inventory (PCI) \cite{pekala_phenomenology_1991}.  Analysis of the 11D-\gls{asc} dimensions revealed predominantly elevated scores in the \textit{Elementary Imagery} and \textit{Complex Imagery} dimensions. The 5D-\gls{asc} results indicated the strongest effects in \textit{Vigilance Reduction} and \textit{Visionary Restructuralization}, with minimal effects on \textit{Oceanic Boundlessness}. The PCI did not reveal any notable alterations in consciousness compared to a control condition, suggesting limited overall alteration in consciousness. These findings suggest that within the laboratory, periodic fixed frequency \gls{sls} primarily induces specific visual phenomena (e.g., simple patterns and colours) and changes in vigilance, rather than broader alterations in consciousness \cite{bartossek_altered_2021, schmidt_altered_2018}.

However, traditional \gls{asc} questionnaires rely on predefined dimensions, which may not fully capture unexpected or idiosyncratic experiences, potentially overlooking less common or individually distinctive aspects of \gls{asc}. To address these limitations and capture the full spectrum of stroboscopically-induced phenomenology without assumptions about the types of experience that may be present, we adopted a data-driven approach. We analysed  862 sentences extracted from 422 free-form open reflections of \gls{sls} experiences collected as part of the Dreamachine programme\footnote{\url{https://dreamachine.world/}}. The Dreamachine is an immersive multisensory experience that uses stroboscopic lighting and 360-degree spatial sound in a highly curated context. It ran in the four capital cities of the UK in 2022, reaching tens of thousands of people, who underwent the experience in collective settings (groups of 20-30 people).

Central to our approach is \gls{tm}, a statistical method for identifying abstract themes in text data \cite{blei_latent_2003,blei_topic_2009}. \gls{tm} detects patterns of co-occurring words,  grouping them into topics that reveal latent semantic structures. This technique offers a more granular exploration of text-based subjective data, capturing elements that traditional structured questionnaire assessments may overlook. By removing the constraints of predefined categories, TM allowed us to explore a wider spectrum of stroboscopically induced phenomenology within the Dreamachine dataset. We hypothesised that this approach would uncover experiential dimensions not previously identified through structured assessment, revealing latent experiential categories within the Dreamachine dataset.

We also make a methodological contribution by implementing and documenting an open-source \gls{nlp}-based pipeline for the analysis of phenomenological free text reports (MOSAIC: Mapping Of Subjective Accounts into Interpreted Clusters)\footnote{\url{https://github.com/romybeaute/MOSAIC}}. Our approach bridges qualitative and quantitative research, providing a systematic, scalable method for analysing rich subjective textual data. To encourage widespread adoption, we provide a fully documented implementation of our analytical workflow, from preprocessing to interpretation. While we apply this approach to stroboscopically induced phenomenology, MOSAIC has broader applications across fields that require systematic analysis of complex and challenging-to-interpret free text reports. We outline the necessary adaptations for different datasets and discuss potential limitations.

\newpage

\section{MATERIALS \& METHODS}
\label{sec:matmethods}

\subsection{Data description}\label{sec:data}

The dataset comprises free text reports collected from participants attending the Dreamachine. The Dreamachine is an installation designed to ``deliver an immersive and multisensory experience that uses flickering white light and 360-degree spatial sound to create a kaleidoscopic and technicolour world behind closed eyes''\footnote{\url{https://dreamachine.world/experience/}}. The data analysed here were collected using a custom tablet application for a digital survey designed, in collaboration with Collective Act, to capture a wide range of participant experiences following the Dreamachine.

To accommodate participants with sensitivity to flashing lights (e.g., photosensitive epilepsy, sensory sensitivity), two versions of the Dreamachine experience were developed. The "\gls{hs}" version consisted of stroboscopic lighting and a 360-degree spatial sound experience. The "\gls{dl}" version offered a parallel experience without the use of stroboscopic lights, instead using sequences of coloured (non-stroboscopic) wash lights. The \gls{dl} Dreamachine employed the same musical soundtrack as the \gls{hs} version, creating an immersive audio environment that encouraged relaxation and inward focus. For the \gls{hs} version, participants were instructed to keep their eyes closed throughout the duration of the experience; for the \gls{dl} version, participants were invited to keep their eyes either open or closed for the duration.

Following their Dreamachine experience, participants were invited to optionally complete the digital survey. This digital survey consisted of two main components: structured questionnaires for specific categories (with some category-specific freeform response options) and a general open-ended reflection. Our study focused specifically on the free-text responses from the general open-ended reflections, where participants provided detailed descriptions of their experiences. The open-ended reflection interface presented participants with a blank space on a tablet, prompted by the following text: ``In Dreamachine I experienced\ldots''.

The study received ethics approval from the University of Sussex (ethics application ID: ER/ANIL/5).

\subsection{Methods}\label{sec:methods}

To identify the most prevalent `dimensions' of experiences present in the data, we employed topic modelling (\gls{tm}), conducting separate analyses on the \gls{hs} and \gls{dl} datasets. \gls{tm} is a \gls{nlp} technique that identifies patterns within text by grouping words and phrases into clusters, or ``topics'', revealing hidden or underlying themes across datasets. These \textit{experiential topics}\footnote{We define `experiential topics' as thematic clusters identified through our TM analysis that represent distinct patterns in participants' subjective reports.} represent recurring themes or patterns in participants' subjective reports, providing a structured method of categorising and understanding the diverse range of experiences.

\gls{tm} can be implemented using various methods, including traditional techniques such as \gls{lda} \cite{blei_latent_2003} and more recently, advanced models based on transformer architectures \cite{thompson_topic_2020, vaswani_attention_2017}. \gls{lda} groups topics by identifying word co-occurrences, but lacks context-awareness -- the ability to interpret how a word's meaning changes based on the context of surrounding words and its position in a sentence -- thus limiting its ability to capture nuanced relationships between words within sentences.

To overcome \gls{lda}'s context limitations, we used an approach combining BERTopic \cite{grootendorst_bertopic_2022} with \gls{llm}-based topic labelling. 
BERTopic addresses these limitations through its transformer-based architecture, enabling context-aware analysis of text data and capturing subtle experiential dimensions by incorporating a contextual understanding of word meanings based on their usage in sentences. 
For instance, whereas \gls{lda} might treat e.g., ``\textit{time}'' as the same way in different contexts (e.g., ``\textit{it was different this time}'' vs. ``\textit{I felt a distortion of time}''), BERTopic's embeddings differentiate generic references for phenomenological implications, thus capturing the multifaceted nature of subjective experiences more accurately.
Finally, we integrated Meta's Llama-3-8B-Instruct \gls{llm} \cite{roziere_code_2023, touvron_llama_2023} to provide automatic, data-driven topic interpretation and labelling based on BERTopic's extracted keywords, rather than relying on subjective manual label generation by the researcher.


\subsubsection{Data pre-processing}\label{sec:preproc}

BERTopic's architectural design offers significant practical and methodological advantages in terms of preprocessing requirements. In practice, this minimises the time and effort needed for data preparation, allowing the model to be applied directly to minimally preprocessed text. BERTopic therefore minimises the potential biases and inconsistencies that can be introduced by preprocessing \cite{rahimi_impact_2023}. These biases and inconsistencies can affect word embeddings and downstream task performance, where word embeddings refer to dense vector representations of words in a high-dimensional space, capturing semantic and syntactic relationships between words.

While BERTopic can handle raw text with minimal preprocessing, our analysis nonetheless required specific preprocessing steps to optimise topic identification. We implemented sentence-level tokenization using a custom \texttt{split\_sentences} function based on the NLTK library \cite{bird_natural_2009}. This methodological choice enables a more granular analysis of experiential reports, allowing each sentence to serve as a distinct unit for embedding and subsequent \gls{tm}, and potentially enhancing the model's ability to capture contextual nuances and semantic meaning \cite{reimers_sentence-bert_2019}. This is particularly valuable for identifying subtle thematic variations within individual reports. Given the multifaceted nature of experience reports, sentence-level tokenization was implemented to enable the differentiation of distinct phenomenological themes within individual reports.

The final count of free text reports and resulting preprocessed sentences are displayed in Table~\ref{tab:dataset_summary}. After preprocessing, the \gls{hs} datasets contained 680 sentences, tokenized from 326 reports. The \gls{dl} dataset contained 182 sentences, tokenized from 96 reports.

\begin{table}[h]
\centering
\begin{tabular}{lcc}
\toprule
\textbf{Dataset} & \textbf{Number of reports} & \textbf{Number of tokenized sentences} \\
\midrule
DL & 96 & 182 \\
HS & 326 & 680 \\
\bottomrule
\end{tabular}
\vspace{1em}
\caption{Summary of the number of reflection answers (reports) per participant, and number of resulting preprocessed sentences after tokenization.}
\label{tab:dataset_summary}
\end{table}

\subsubsection{BERTopic}\label{sec:bertopic}
BERTopic's implementation centres on pre-trained transformer-based language models that generate text embeddings -- high-dimensional vectors that encapsulate the semantic nuances of words, sentences, or entire documents. 
The architecture of BERTopic comprises four key components: (i) a transformer embedding model; (ii) \gls{umap} dimensionality reduction \cite{mcinnes_umap_2018}; (iii) \gls{hdbscan} \cite{mcinnes_hdbscan_2017} clustering; (iv) cluster tagging using class-based Term Frequency-Inverse Document Frequency (c-TF-IDF). This combination produces dense clusters that are easily interpretable and coherent, while retaining important words in the topic descriptions. We detail each of these core components in the following sections.

\paragraph{Transformer embeddings}\label{sec:transfembs}
\hspace{1cm}

The embedding process involved encoding text from the open reports into high-dimensional vector representations using a pre-trained Sentence Transformer model \cite{vaswani_attention_2017}. We selected the ``all-mpnet-base-v2'' model from the sentence-transformers library of the Hugging Face Hub\footnote{\url{https://huggingface.co/sentence-transformers/all-mpnet-base-v2}}. This model was chosen due to its intended use as a sentence encoder, and its strong performance on semantic similarity tasks according to the Massive Text Embedding Benchmark (MTEB) leaderboard\footnote{\url{https://huggingface.co/spaces/mteb/leaderboard}} \cite{muennighoff2022mteb}. This model maps text to a 768-dimensional vector space, providing a rich semantic representation while maintaining computational efficiency.

\paragraph{Dimensionality reduction}\label{sec:dimred}
\hspace{1cm}

After obtaining the high (768) dimensional embeddings, we performed a dimensionality reduction using \gls{umap} python library\footnote{\url{https://pypi.org/project/umap-learn/}}\cite{mcinnes_umap_2018}. \gls{umap} was chosen over alternatives such as t-SNE (t-distributed Stochastic Neighbour Embedding) for its ability to preserve both local and global data structure, which is crucial to maintain semantic relationships in the reduced dimensional space. We optimised \gls{umap} parameters through a grid-search, focusing on three key hyperparameters: \texttt{n\_components} controlling the dimensionality of the reduced space, \texttt{n\_neighbors} balancing local and global structure preservation, and \texttt{min\_dist} determining the minimum distance between points in the embedded space.

\paragraph{HDBSCAN Clustering}\label{sec:HDBSCAN}
\hspace{1cm}

Following dimensionality reduction, we employed \gls{hdbscan} for cluster analysis \cite{mcinnes_hdbscan_2017}. \gls{hdbscan} was selected for its ability to identify clusters of varying densities and shapes without requiring a pre-specified number of clusters. The algorithm automatically determines the optimal number of clusters based on the density structure of the data, making it particularly suitable for exploratory analysis of experiential reports. \gls{hdbscan} parameters were systematically optimised alongside \gls{umap} parameters through an extensive grid search. The search explored combinations of \texttt{min\_cluster\_size} and \texttt{min\_samples}, which control the minimum cluster size and noise tolerance respectively. Parameter combinations were evaluated using topic coherence metrics C\textsubscript{v} and UMass \cite{roder_exploring_2015} to identify the optimal configuration that produced the most interpretable and cohesive topics (see below).

\paragraph{Topic extraction and clustering evaluation}\label{sec:clusteval}
\hspace{1cm}

We evaluated model performance using C\textsubscript{v} and UMass coherence metrics, implemented via the Gensim library \cite{lau_machine_2014,mimno_optimizing_2011,newman_automatic_2010}. C\textsubscript{v} employs sliding window segmentation with normalised pointwise mutual information and cosine similarity, whilst UMass uses sentences co-occurrence counts and logarithmic conditional probability. These metrics assess the meaningfulness of relationships between words within topics, with C\textsubscript{v} in particular showing strong correlation with human judgments of topic quality \cite{roder_exploring_2015}.

Our evaluation strategy optimised both topic coherence and clustering stability, with emphasis on coherence. Parameter optimisation was conducted through extensive grid search, guided by heuristics and initial exploratory results. While coherence scores served as a primary metric for parameter selection, we also considered the number of topics extracted from each parameter combination to achieve a balance. Specifically, we filtered out combinations that led to too few topics ($<$5) to avoid overly broad and non-informative categories, and too many topics ($>$25), which otherwise lead to overfitting, fragmented, and non-informative topic extraction, based on practical considerations of interpretability and the size of our dataset. The optimisation process yielded different parameter sets for each condition (Table~\ref{tab:hyperparams}), with the highest coherence scores of 0.56 (\gls{hs}, 14 topics) and 0.57 (\gls{dl}, 8 topics), reflecting dataset-specific characteristics. These coherence scores fall within the acceptable range typically observed in \gls{tm} studies (0.4--0.7)\cite{ocallaghan_analysis_2015,roder_exploring_2015}.

\begin{table}[h]
\centering
\begin{tabular}{lccccccc}
\toprule
& \texttt{n\_comp} & \texttt{n\_neigh} & \texttt{min\_dist} & \texttt{min\_clust} & \texttt{min\_samp} & \texttt{coh\_score} & \texttt{n\_topics} \\
\midrule
HS & 15 & 20 & 0.025 & 10 & 5 & 0.56 & 14 \\
DL & 18 & 25 & 0.025 & 5 & 5 & 0.57 & 8 \\
\bottomrule
\end{tabular}
\vspace{0.5em}
\begin{flushleft}

\caption{Results of a grid-search for the two versions of the Dreamachine: High Sensory (HS) and Deep Listening (DL).}
\label{tab:hyperparams}
\end{flushleft}
\end{table}

{\textbf{LLM-based topic labelling}}\label{sec:finetuning}
\hspace{1cm}

After BERTopic identified topic clusters, we used a large language model (Llama-3-8B-Instruct) \cite{roziere_code_2023,touvron_llama_2023} to automatically generate descriptive labels for each topic, enhancing topic interpretability. Specifically, we provided the \gls{llm} with the top c-TF-IDF keywords for each cluster (typically 10-15 keywords) along with short representative text excerpts from the same cluster. This approach was intended to shift the labelling process from a subjective manual task to a data-driven method implemented by the \gls{llm}. Since \glspl{llm} are generative models, some inherent stochasticity exists, meaning that repeated labelling attempts multiple times may produce slight variations in phrasing (e.g., “Imagery” vs. “Visuals”). However, because the same c-TF-IDF keywords and representative snippets were used each time to generate the labels, the core semantic theme remained stable across runs.  To verify label reliability, we conducted multiple labelling attempts and confirmed that variations were limited to wording rather than underlying conceptual meaning.
To implement this step, we used a quantified version of Llama-3-8b-instruct, ``Meta-Llama-3-8B-Instruct-Q4\_K\_M.gguf''\footnote{\url{https://huggingface.co/meta-llama/Meta-Llama-3-8B-Instruct}} along with llama-cpp-python implementation from Hugging Face\footnote{\url{https://huggingface.co/docs/chat-ui/en/configuration/models/providers/llamacpp}}.

\hspace{20cm}

We used Python to implement all data analysis, \gls{tm}, and visualisations. \gls{tm} was primarily conducted using the BERTopic library (version 0.16.2), leveraging \gls{umap} for dimensionality reduction and \gls{hdbscan} for clustering. For sentence embeddings, we used the SentenceTransformer model (sentence-transformers: all-mpnet-base-v2). Data handling and manipulation were performed with pandas (v2.0.3) and NumPy (v1.24.4). Visualisation of the generated topics and embeddings were created using custom scripts, with Matplotlib and Seaborn employed for \gls{lda} and hierarchical clustering visualisation. We used a grid search to optimise hyperparameters, fine-tuning the topic model to accurately capture the diversity of the dataset, allowing for more effective visualisation and interpretation of topic representations for both the \gls{hs} and \gls{dl} datasets.

\begin{figure}[htbp]
    \centering
    \includegraphics[width=0.95\textwidth]{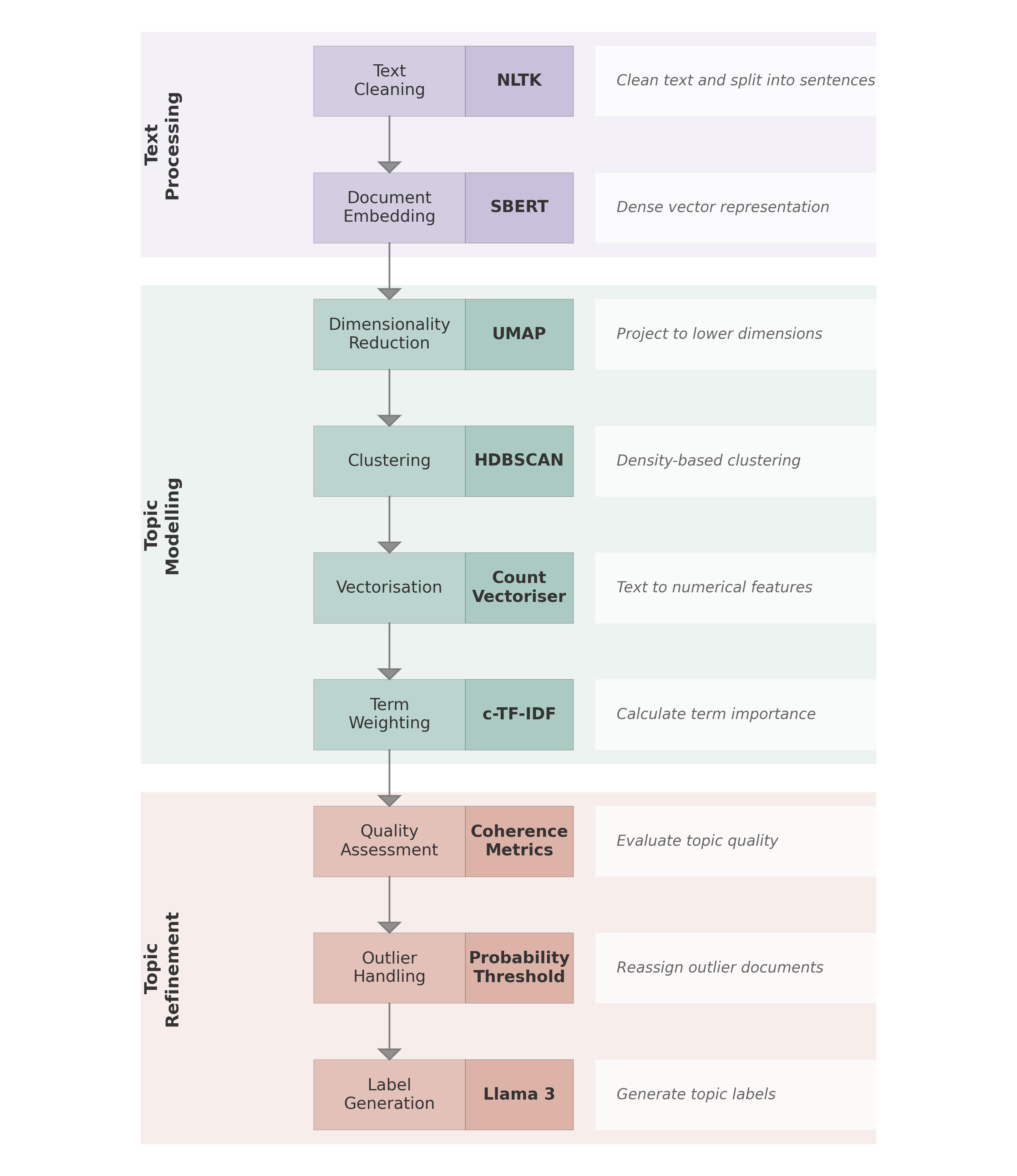}  
    \caption{Topic modelling pipeline architecture comprising three phases: text processing (NLTK for sentence segmentation, SBERT for dense embeddings), topic modelling (UMAP for dimensionality reduction, HDBSCAN for clustering, Count Vectorisation and c-TF-IDF for term weighting), and topic refinement (Coherence Metrics for quality assessment, Probability Threshold for outlier handling, Llama 3 for label generation)}
    \label{fig:topmod_schema}
\end{figure}

\newpage
\section{RESULTS}

\subsection{High Sensory (HS) dataset: topic modelling and representation}

\begin{figure}[h!]
    \centering
    \includegraphics[width=1\textwidth]{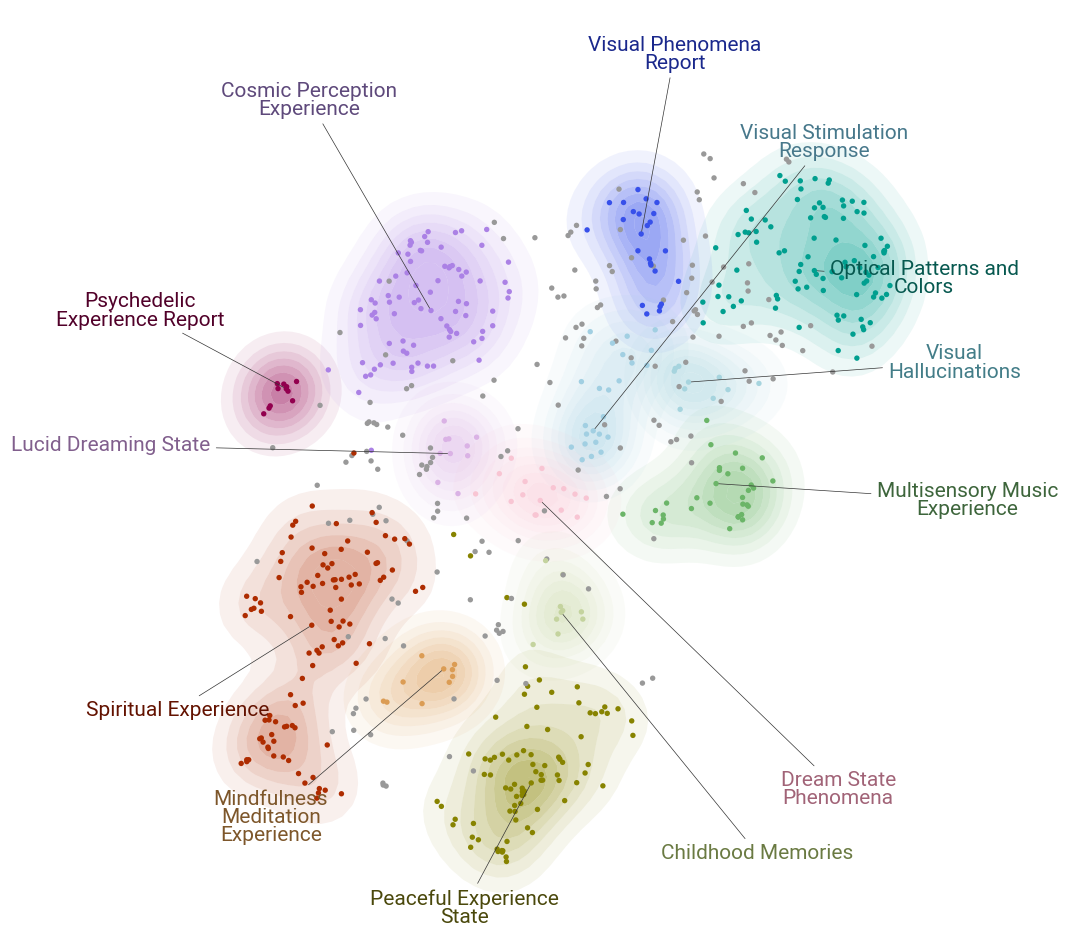}  
    \caption{\textbf{Topic Representations of HS Dreamachine experiences}\\
Two-dimensional embedding visualisation of experiential topics (n=13) derived from HS Dreamachine participant reflections (n=680 sentences). Each point represents a sentence, colour-coded by its dominant topic. Spatial proximity indicates semantic similarity between reports, computed using BERTopic’s transformer-based embeddings. The visualisation reveals distinct phenomenological domains: visual-perceptual (turquoise-blue), altered states of consciousness (purple), and emotional-meditative experiences (red-orange), and multisensory/autobiographical experiences (green). Topic labels were generated using Llama-3-8b-instruct, which interpreted the underlying semantic clusters. Points are clustered by semantic similarity, with overlapping regions suggesting shared phenomenological features.
}\label{fig:HSfig}
\end{figure}

\begin{figure}[h!]
    \centering
    \includegraphics[width=1\textwidth]{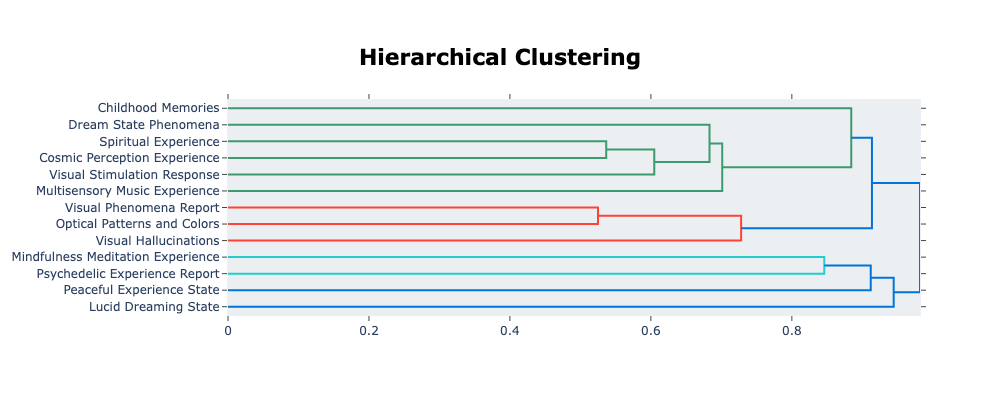}  
    \caption{\textbf{Hierarchical clustering of HS Dreamachine experiences}\\
Dendrogram showing semantic relationships between 13 topics identified from HS Dreamachine reflections. Topics were extracted using BERTopic and labelled with Llama-3-8b-instruct. Vertical distances represent cosine dissimilarity between topics. The clustering reveals distinct phenomenological domains: autobiographical-spiritual experiences (green cluster: including childhood memories, spiritual experiences, and multisensory experiences), visual phenomena (red cluster: optical patterns, visual hallucinations) and altered states experiences (blue-turquoise cluster: psychedelic experiences, mindfulness meditation, peaceful states, lucid dreaming). This hierarchical view complements the spatial relationships shown in Figure \ref{fig:HSfig}
\label{fig:HS_HC}}
\end{figure}

\gls{tm} of participants' \gls{hs} Dreamachine reflections revealed thirteen distinct experiential topics, after removing outliers (which in BERTopic are designated as 'topic -1' and represent sentences that could not be confidently assigned to any cluster). These topics were labelled by Llama 3 automatically based on the c-TF-IDF keywords and representative text for each cluster, ensuring minimal researcher bias in the labelling. Figures \ref{fig:HSfig} and \ref{fig:HS_HC} display the relationships among these topics through 2D embedding visualisation and hierarchical clustering, respectively. Spatial proximity in Figure \ref{fig:HSfig} indicates thematic similarity, and the dendrogram (Figure \ref{fig:HS_HC}) quantifies relationships through cosine distances (0-1, lower values indicating stronger connections).

In the spatial embedding (Figure \ref{fig:HSfig}), the distribution of topics reveals several phenomenological domains: (i) visual-phenomena (turquoise-blue region) encompassing \textit{Optical Patterns and Colors}, \textit{Visual Stimulation Responses,} and \textit{Visual Hallucinations}; (ii) altered states including \textit{Psychedelic Experience Report} and \textit{Lucid Dreaming State} (purple region); (iii) spiritual and meditative, and peaceful states (red-orange region); and (iv) autobiographical elements including \textit{Childhood Memories} (green region). The proximity between points suggests semantic relationships between these experiential aspects, with some overlap between domains potentially indicating shared phenomenological features.

Analysis revealed thirteen (automatically labelled) distinct experiential topics:

\begin{enumerate}
    \item \textbf{Spiritual Experience} (derived from n=113 sentences), characterised by profound emotional states and feelings of connection and awe: \textit{``I felt an overwhelming sense of euphoria and love for everyone in the room''} and \textit{``The experience felt almost transformative''}.
    
    \item \textbf{Optical Patterns and Colours} (derived from n=99 sentences) features detailed geometric and kaleidoscopic effects: \textit{``intense geometric shapes spiraling into a deep ocean blue circle''} \textit{``shockingly vivid white and red and black, angular kaleidoscope with 3D elements''}.
    
    \item \textbf{Cosmic Perception Experience} (derived from n=99 sentences) describes space-like journeys and altered perceptions: \textit{``flying through space at incredible speed''} \textit{``I felt like I was looking across the Universe beyond any realm of experience''}.
    
    \item \textbf{Peaceful Experience State} (derived from n=90 sentences) captures states of deep relaxation and tranquility: \textit{``an amazing sense of harmony, balance, and oneness, visually a moving mandala''} \textit{``Connection to an inner peace''}.
    
    \item \textbf{Visual Phenomena Report} (derived from n=42 sentences) captures specific visual content such as \textit{``geometric kaleidoscope shapes overlaid with moving cosmos/night sky''} \textit{``sea of green stars moving like oil on water, wading fading''}.
    
    \item \textbf{Multisensory Music Experience} (derived from n=37 sentences) highlights audio-visual integration: \textit{``Sound created the visuals and created a pattern of energy movement across my body''} \textit{``I became the music and the music became me, and we were floating and flying in a multidimensional kaleidoscope''}.
    
    \item \textbf{Visual Stimulation Response} (derived from n=28 sentences) seems to relate more to physiological impacts of the light: \textit{``Bright light was a bit hurting but overall experience was amazing''} \textit{``I could feel some sort of physical effect/stress or pressure on the front of my brain''}.
    
    \item \textbf{Dream State Phenomena} (derived from n=26 sentences) seems to capture subconscious states: \textit{``And then my mind checked out and my subconscious took over''}, \textit{``I found my thoughts developing into stories and images while my eyes were closed''}.
    
    \item \textbf{Psychedelic Experience Report} (derived from n=11 sentences) drew comparisons to previous exposure to psychedelics: \textit{``closest sober experience (and to be honest, better) than psychedelics''} \textit{``this is exactly how I remember being on LSD was, without the scary parts''}.
    
    \item \textbf{Visual Hallucinations} (derived from n=21 sentences) describes hallucinations with a strong influence of visual aspects of such experience, such as \textit{``lots of visuals, but towards the end people I love, and the ancestors''} and \textit{"Colors, patterns, and a sense of awe that my brain and light were creating such powerful hallucinations."}.
    
    \item \textbf{Childhood Memories} (derived from n=14 sentences) captures autobiographical recollections: \textit{``The most vivid memories of my Dad carrying me on his shoulders''}, \textit{``I remembered beautiful and joyful memories and moments with my family''}
    
    \item \textbf{Lucid Dreaming State} (derived from n=13 sentences) describes dream-like experiences such as \textit{``dreaming while awake- flashes of random places I have been''}, \textit{``I felt like I drifted in and out of consciousness, but never opened my eyes''}
    
    \item \textbf{Mindfulness Meditation Experience} (n=12 sentences) captures reports related to meditative states: \textit{``This would be a great therapy to teach mindfulness''} \textit{``the meditative part was nice''}
\end{enumerate}

The hierarchical clustering analysis (Figure \ref{fig:HS_HC}) reveals distinct phenomenological groupings. Essentially, three major clusters emerged: (i) a visual experience cluster (cosine distance 0.6--0.8) comprising \textit{Visual Phenomena Report}, \textit{Optical Patterns and Colors}, and \textit{Visual Hallucinations}, highlighting the prominence of visual elements in participants' experiences; (ii) an altered states cluster (cosine distance 0.8--1.0) including \textit{Psychedelic Experience Report}, \textit{Peaceful Experience State}, and \textit{Lucid Dreaming State}, suggesting common experiential qualities between these consciousness-altering states; and (iii) a memory-spiritual cluster (cosine distance 0.2--0.4) connecting \textit{Childhood Memories}, \textit{Dream State Phenomena}, and \textit{Spiritual Experience}.

The distribution of sentences across topics shows considerable variation. Dominant topics ($>$80 sentences) included \textit{Spiritual Experience} (derived from n=104 sentences), \textit{Optical Patterns and Colors} (derived from n=92 sentences), \textit{Peaceful Experience State} (derived from n=85 sentences), and \textit{Cosmic Perception Experience} (derived from n=84 sentences), suggesting these were core elements of the Dreamachine experience. Moderate-frequency topics (25--80 sentences) comprised \textit{Multisensory Music Experience} (derived from n=32 sentences) and \textit{Visual Phenomena Report} (derived from n=25 sentences), while low-frequency topics ($<$25 sentences) included \textit{Visual Stimulation Response} (derived from n=24 sentences), \textit{Dream State Phenomena} (derived from n=16 sentences), and several topics with approximately 10 sentences each (\textit{Psychedelic Experience, Visual Hallucinations, Childhood Memories, Lucid Dreaming, and Mindfulness Meditation Experience}).

\begin{table}[h]
\centering
\begin{tabular}{lrr}
\toprule
\textbf{Topic Label} & \textbf{Sentence Count} & \textbf{Percentage (\%)} \\
\midrule
Spiritual Experience & 113 & 16.62 \\
Optical Patterns and Colors & 99 & 14.56 \\
Cosmic Perception Experience & 99 & 14.56 \\
Peaceful Experience State & 90 & 13.24 \\
Unlabelled & 75 & 11.03 \\
Visual Phenomena Report & 42 & 6.18 \\
Multisensory Music Experience & 37 & 5.44 \\
Visual Stimulation Response & 28 & 4.12 \\
Dream State Phenomena & 26 & 3.82 \\
Visual Hallucinations & 21 & 3.09 \\
Childhood Memories & 14 & 2.06 \\
Lucid Dreaming State & 13 & 1.91 \\
Mindfulness Meditation Experience & 12 & 1.76 \\
Psychedelic Experience Report & 11 & 1.62 \\
\midrule
Total & 680 & 100.00 \\
\bottomrule
\end{tabular}
\vspace{1em}
\caption{Distribution of experiential topics in High Sensory (HS) Dreamachine reports.}
\label{tab:HS_topics}
\end{table}

\subsection{Deep Listening (DL) dataset: topic modelling and representation}

The \gls{tm} analysis of participants' reflections following the \gls{dl} Dreamachine yielded seven distinct topics, capturing a range of sensory, cognitive, and emotional experiential dimensions. The topics were identified by BERTopic and automatically labelled by Llama 3. 

Comparing label assignments between the HS and DL datasets, there are both differences (e.g., "\textit{Music perception}" appears in DL with no obvious complement in HS, and vice-versa for "\textit{Optical patterns and colors}") and similarities (e.g. "\textit{Dream state phenomena}" in HS and "\textit{Dream imagery}" in DL).  Differences in topic labels may reflect genuine variations in the nature of participants experiences across these two versions of the Dreamachine (though some minor differences may result from the stochastic, generative nature of Llama 3).

\begin{figure}[h!]
    \centering
    \includegraphics[width=1\textwidth]{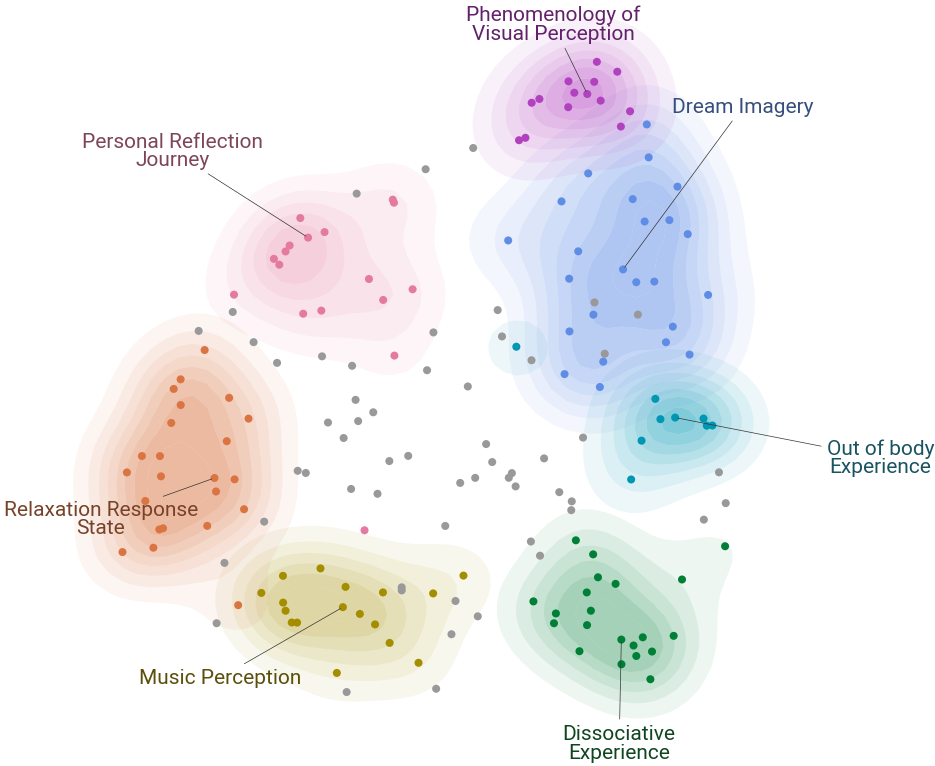}  
    \caption{\textbf{Topic Representations of DL Dreamachine experiences}\\
Two-dimensional embedding visualisation of experiential topics (n=7) derived from DL Dreamachine participant reflections (n=182 sentences). Each point represents a sentence, colour-coded by its dominant topic. Spatial proximity indicates semantic similarity between reports, computed using BERTopic's transformer-based embeddings. The visualisation reveals distinct experiential domains: sensory-perceptual (purple), mental imagery (red-orange), mindfulness states (green-blue), and out-of-body experiences (grey). Topic labels were generated using Llama-3-8b-instruct, which interpreted the underlying semantic clusters. Points are clustered by semantic similarity, with overlapping regions suggesting shared phenomenological features.
}\label{fig:DLfig}
\end{figure}

\begin{figure}[h!]
    \centering
    \includegraphics[width=1\textwidth]{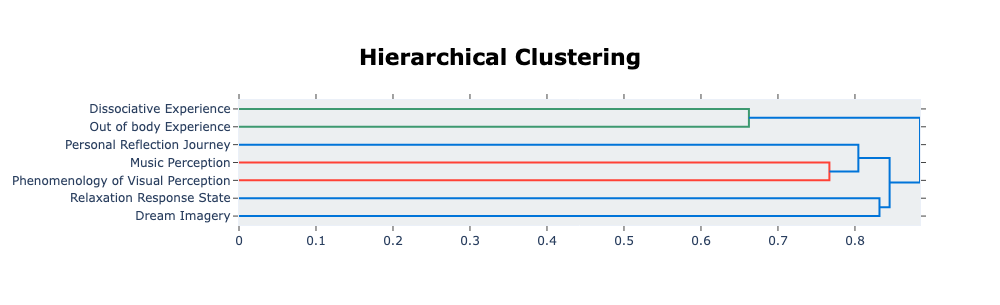}  
    \caption{\textbf{Hierarchical clustering of DL Dreamachine experiences}\\
Dendrogram showing semantic relationships between 7 topics identified from DL Dreamachine reflections. Topics were extracted using BERTopic and labelled with Llama-3-8b-instruct.
\label{fig:DL_HC}}
\end{figure}

Conversely, similar topic labels do not necessarily indicate similar experiences, as participants may use overlapping language to describe qualitatively distinct states. For example, a “dream-like state” could emerge both from listening to music with eyes closed and from the combination of music and \gls{sls}, yet these experiences may differ significantly in their underlying phenomenology. The resulting labels from their top keywords are displayed in Figure \ref{fig:DLfig}. Figure \ref{fig:DL_HC} illustrates the hierarchical clustering of these topics, demonstrating their relationships and similarities.

The two-dimensional embedding visualisation (Figure \ref{fig:DLfig}) reveals topical clustering including \textit{Phenomenology of Visual Perception} (purple region), \textit{Dream Imagery} (blue region), \textit{Personal Reflection Journey} (pink region), \textit{Relaxation Response State} (orange region), \textit{Music Perception} (yellow region), \textit{Dissociative Experiences} (green region), as well as \textit{Out-of-body Experiences} (turquoise region).

The \gls{tm} analysis of the \gls{dl} Dreamachine revealed seven (automatically labelled) distinct experiential topics, characterised by the following themes and representative quotes:

\begin{enumerate}
    \item \textbf{Dream Imagery} (derived from n=24 sentences) contains vivid descriptions of mental imagery and hallucination-like experiences. Participants reported complex scenes, such as \textit{``I went on a train, underwater, saw seas of colour and pulsing shapes''}, \textit{``I was in a land (not anywhere on Earth) it was another planet, with animals, but they were not the usual colours - a yellow elephant, a blue gorilla''}, \textit{``Got many vivid visions, saw lots of colours, my bedroom, old TV shows, video games, cities\ldots''}.
    
    \item \textbf{Personal Reflection Journey} (derived from n=17 sentences) captures introspective experiences and autobiographical experiences. Participants described self-referential thoughts, such as \textit{``Looking to the past to analyse my future''} and \textit{``I saw past experiences and reflected on my choices''}.
    
    \item \textbf{Phenomenology of Visual Perception} (derived from n=14 sentences) focuses on more direct sensory experiences, particularly the interplay between external stimuli and internal perception: \textit{``The colours behind my closed eyes swapped to the opposite colour or the colour spectrum of that on the screen''} and \textit{``the colours i saw when my eyes were shut were very brief, fleeting and hues, nebulous, no specific patterns''}, \textit{``Interested to experience different darkness when different colours were projected''}.
    
    \item \textbf{Music Perception} (derived from n=17 sentences) reveals reports of alterations in auditory experiences and its integration with other senses. Participants reported complete immersion: \textit{``Music wasn't just in the space, it was the space''}, \textit{``physically being the music, especially the percussion and the bass''}, \textit{``Thick hot air saturated with the soundwaves which I was inhaling''}. Some descriptions linked music to visual imagery: \textit{``my visual experience would change a lot with the music''} and \textit{``I watched the room rise and fall with the music''}.
    
    \item \textbf{Dissociative Experience} (derived from n=21 sentences) indicates reports of alterations in bodily awareness. Participants described sensations of descending or sinking: \textit{``felt as though I was moving through a tunnel at first''}, \textit{``I felt like my body kept sinking or dropping to another level... into the seat and into itself''}. Several reports emphasised oceanic or aquatic experiences: \textit{``It felt like I was deep in the ocean''} and \textit{``Almost womb like... And it also felt like being inside an ocean''}. Others described feelings of anxiety and physical discomfort during these dissociative states: \textit{``I felt intensely anxious and a bit suffocated but also knowing that I've got this''} and \textit{``First drowning in that sound, then learning how to swim and enjoy it and heal''}. 
    
    \item \textbf{Relaxation Response State} (derived from n=23 sentences) captures experiences of deep calm and meditative states. Participants reported profound relaxation and peace: \textit{``Complete relaxation, peace of mind and ready to face whatever happens''}, \textit{``Vibration in visuals Centered in body Aware of sounds Mind really quietened down''}. They described states of deep tranquility: \textit{``Peacefulness and calm''} and \textit{``I experienced a calmness almost a nothingness but it was relaxed and peaceful and soothing''}.
    
    \item \textbf{Out of body Experience} (derived from n=9 sentences) seems to capture instances of mental detachment and memory recall, distinct from physical dissociation. Participants reported dream-like detachment from physical reality: \textit{``some thoughts about mundane everyday things like work, but mainly felt relaxed and able to float into a dreamlike place some visuals of space and deep ocean''} and altered time perception: \textit{``i can't realise how long was that experience it was so quick in and then out i don't feel the time''}, suggesting a distinct phenomenological state focused on mental rather than bodily experience.
\end{enumerate}

A substantial portion of responses (n=57, 31.32\%) remained unassigned by the \gls{tm} analysis. These unassigned responses comprised various types of commentary: feedback about the physical environment (e.g., \textit{it was a bit too cold for me''}), reflections on session quality (e.g., \textit{this time someone kept coughing and disrupted my thought process''}), hypotheses about effects (e.g., \textit{``i think i ought to experience it once more to see if it changes''}), and methodological suggestions (e.g., \textit{``thus should be available for everyone all the time''}). Some responses contained declarative statements about future engagement that did not align with the identified experiential clusters (e.g., \textit{``I want one of these at home''} or \textit{``I want to do it again!''}). The high proportion of unassigned responses indicates considerable heterogeneity in participants' reports, with some responses focusing on procedural aspects, rather than descriptions of their inner experiential content.

\begin{table}[h!]
\centering
\begin{tabular}{lrr}
\toprule
\textbf{Topic Label} & \textbf{Sentence Count} & \textbf{Percentage (\%)} \\
\midrule
Unlabelled & 57 & 31.32 \\
Dream Imagery & 24 & 13.19 \\
Relaxation Response State & 23 & 12.64 \\
Dissociative Experience & 21 & 11.54 \\
Music Perception & 17 & 9.34 \\
Personal Reflection Journey & 17 & 9.34 \\
Phenomenology of Visual Perception & 14 & 7.69 \\
Out of body Experience & 9 & 4.95 \\
\midrule
Total & 182 & 100.00 \\
\bottomrule
\end{tabular}
\vspace{1em}
\caption{Distribution of experiential topics in Deep Listening (DL) Dreamachine reports.}
\label{tab:DL_topics}
\end{table}

\FloatBarrier
\FloatBarrier

The hierarchical clustering analysis (Figure \ref{fig:DL_HC}) revealed meaningful relationships between these experiential domains. Most notably, dissociative and out-of-body experiences showed the closest semantic similarity, whilst visual perception and relaxation states formed a separate major branch. This clustering suggests at least two primary dimensions of experience: altered self-experience and sensory-perceptual modulation.

\section{DISCUSSION}
To advance our understanding of stroboscopically-induced phenomenology, beyond traditional categorisations, we employed a data-driven approach combining \gls{tm} and \gls{llm} to analyse participants’ unstructured free text reports from the Dreamachine multisensory immersive experience. This approach addressed the limitations of structured questionnaires, which may constrain or bias the capture of diverse phenomenological experiences. It also leveraged the opportunity presented by the novel Dreamachine dataset. Our analysis revealed that stroboscopically-induced phenomenology encompasses a broader spectrum of experiences than previously documented, identifying three principal categories: simple \glspl{vh}, complex \glspl{vh}, and \glspl{asc}.

\subsection{Phenomenology of the Dreamachine}

\subsubsection{Simple Visual Hallucinations (VHs)}
Simple \glspl{vh} encompass a range of basic \glspl{vh} characterised by their lack of narrative  content. These phenomena include phosphenes, geometric forms, vivid colours, and fractal or kaleidoscopic patterns \cite{bressloff_what_2002,rule_model_2011,shevelev_visual_2000,smythies_stroboscopic_1959}. Their classification as “simple” stems from their abstract nature rather than visual complexity (e.g. fractals). 

The prevalence of simple \glspl{vh} in the \gls{hs} version of Dreamachine is consistent with the established literature on \gls{sls}-induced phenomena \cite{allefeld_flicker-light_2011,billock_elementary_2012,young_subjective_1975}. Participants frequently reported experiencing vibrant geometric patterns, kaleidoscopic visuals, and dynamic colour phenomena. These experiences were primarily captured in topics such as “Optical Patterns and Colours”, which focused on basic geometric forms and colour perception (e.g., \textit{“shockingly vivid white and red and black, angular kaleidoscope with 3D elements”}) and “Visual Perception Phenomena”, which emphasised the dynamic aspects of the visual experiences (e.g., \textit{“geometric shapes shifting and changing”}). 

These experiences likely originate from altered perceptual processing in lower-level visual areas, particularly the primary visual cortex (V1) \cite{bressloff_what_2002,kometer_serotonergic_2018}. Neural field models posit that simple \glspl{vh} should be organised around the centre of the visual field, where the highest level of detail occurs, as exemplified by the tunnel, spiral, and target Klüver forms \cite{bressloff_what_2002}. These models align with the architecture and function of V1, where neurons are organised into hypercolumns responding to specific orientations and spatial frequencies \cite{rule_model_2011}. The architecture of V1 is therefore believed to be a good candidate underlying the perception of the specific reported geometric properties of simple \glspl{vh}, such as grids, lattices, and spirals. \gls{sls} may lead groups of neurons in V1 to fire in synchrony, inducing resonance in these neural populations, leading to the perception of geometric hallucinations \cite{billock_elementary_2012}. This resonance effect could explain the prevalence of certain geometric forms in stroboscopically-induced simple \glspl{vh}, with the most commonly reported forms likely corresponding to the most stable or easily induced patterns of neural activity in V1.

\subsubsection{Complex Visual Hallucinations}
Complex \glspl{vh} involves the perception of dream-like hallucinations of meaningful or recognisable percepts, such as realistic scenes, objects, or faces. Differing from simple \glspl{vh} described above, we consider hallucinations as ``complex'' when they involve higher-level cognitive processing, often incorporating elements from memory, imagination, or narrative features. A notable finding of our study is the frequent reports of complex \glspl{vh}, particularly evident in the ``Visual Hallucinations'' topic. Participants reported seeing semantically rich content such as \textit{``a deep red heart in the distance as well as flowers, glitter, skies, an eye and an owl''} and complex scenes like \textit{``A wheat field, A girl in summer, A roller coaster speeding, A bird flying out of the earth's atmosphere and into space''}. In some cases, these complex visual experiences seemed to incorporate personally meaningful elements, e.g. \textit{``Seeing people I love, and the ancestors... within an eye''}. These complex \glspl{vh} reports often combined abstract patterns with recognisable forms, suggesting the engagement of higher-order visual and cognitive processing. 

Several factors may contribute to the generation of complex \glspl{vh} during \gls{sls}. From a neural perspective, \gls{sls} may disrupt the balance of excitatory and inhibitory activity in early visual cortices, subsequently triggering downstream activation of higher-order visual areas including the ventral temporal cortex and associative regions. This pattern of hierarchical activation could explain the progression from simple geometric patterns to more complex, meaningful percepts. This distributed activation likely involves regions supporting memory retrieval (hippocampus), emotional processing (amygdala), and cognitive control (prefrontal cortex). The interaction between these neural systems may explain why complex \glspl{vh} often incorporate personally meaningful content and emotionally-coloured imagery. Another simple explanation for the occurrence of complex \glspl{vh} is that with eyes closed, participants who reported complex \glspl{vh} may have entered a drowsy hypnagogic state, which is commonly associated with highly vivid visual experiences lacking a narrative or overarching themes \cite{wackermann_ganzfeld-induced_2008,ghibellini_hypnagogic_2023}. These neural and cognitive processes may be further modulated by top-down influences, particularly through predictive processing mechanisms where prior expectations shape perceptual inference \cite{powers_hallucinations_2016}.

Another possibility is that individual differences in visual mental imagery, or the ability to create the experience of viewing a stimulus in the absence of appropriate sensory input, contributes to the generation of complex hallucinations. Participants who score high in visual mental imagery may be more likely to experience complex \glspl{vh} as a result of projecting their highly realistic visual imagery into their stroboscopic experience. 

Other psychological traits related to hypnotic suggestibility may also play a part, such as `absorption' - which relates to the tendency to become immersed in experiences \cite{bartossek_altered_2021} - and `phenomenological control' \cite{lush_phenomenological_2021}, which refers to an individual's ability to alter what they experience, both within and outside of the hypnotic context, in ways that are consistent with their intentions and goals. 
These traits may exert their effects through participants' implicit or explicit expectations - `demand characteristics' - about the Dreamachine experience. Trait phenomenological control scores are known to predict subjective responses in various experimental contexts where demand characteristics are evident, such as the rubber hand illusion and Mirror-touch synaesthesia \cite{lush_trait_2020}. More recently, they have been linked to auditory `hallucinations' such as those associated with silent moving images of objects colliding (visually evoked auditory response (vEAR)) \cite{lush_phenomenological_2024}. Most of the participants in the Dreamachine were aware that they might experience \glspl{vh}, which likely presented demand characteristics that shaped their experiences and survey responses in ways that align with the effects of phenomenological control. Future studies should test this hypothesis by screening participants, e.g., using the Phenomenological Control Scale (PCS) \cite{lush_phenomenological_2021}, to examine individual differences in phenomenological control and their potential predictive role in the occurrence of both complex and simple \glspl{vh}, along with other commonly reported stroboscopic altered states phenomena.

\subsubsection{Altered States of Consciousness}

\gls{asc} experiences describe mental states that differ profoundly from typical, everyday states of consciousness. These experiences may include alterations in how the external world is perceived, variations in the awareness of one's emotions, sensations, and thoughts, a distorted sense of space and time, changes in self-awareness including ego dissolution, or feelings of unity or oneness with the environment \cite{dittrich_standardized_1998,schmidt_altered_2018,studerus_psychometric_2010}. 

We observed, unexpectedly, \gls{asc} reports in both Dreamachine versions, although the number of explicit \gls{asc} reports was much lower in the \gls{dl} compared to the \gls{hs} version. Topics from the \gls{hs} dataset captured a range of altered states, from psychedelic-like experiences, illustrated by the ``Psychedelic Experience Report'' topic (e.g. \textit{``this is exactly how I remember being on LSD was''}), to transitions in consciousness states in ``Dream State Transition'' (e.g. \textit{``I felt like I drifted in and out of consciousness''}), meditative states in ``Mindfulness Meditation Experience'' (e.g \textit{``like a deep tissue massage for the mind''}) and profound states of calm and unity captured in ``Peaceful Experience State'' (e.g. \textit{``an amazing sense of harmony, balance, and oneness''}).

\gls{asc} reports like these have not typically been observed in laboratory \gls{sls} studies, to our knowledge \cite{bartossek_altered_2021}. While potentially sharing some neural mechanisms with complex \glspl{vh}, \glspl{asc} may represent a broader alteration of consciousness that extends beyond visual processing. The naturalistic setting of the Dreamachine, compared to laboratory \gls{sls} studies, may create different expectational contexts that influence both complex \glspl{vh} and \glspl{asc}, though potentially in distinct ways. For instance, while complex \gls{vh} expectations might focus on specific visual content, \gls{asc} expectations may relate more to overall state changes. These different types of expectations could engage distinct neural and cognitive mechanisms, despite potentially overlapping neural substrates \cite{howard_anatomy_1998,ffytche_hodology_2008,kometer_activation_2013,wittmann_effects_2007}. The interaction between bottom-up and top-down processes in generating these experiences may depend on various factors, including expectations, naturalistic setup, and neural entrainment \cite{amaya_effect_2023,carhart-harris_rebus_2019,carhart-harris_psychedelics_2018,schwartzman_increased_2019}. Future studies should systematically investigate how these contextual factors and individual differences influence the emergence of both complex \glspl{vh} and \glspl{asc} during \gls{sls}.

Although some topic labels in the \gls{dl} dataset (e.g., "\textit{Out of Body Experiences"}) might suggest altered states, these generated labels do not necessarily indicate that participants experienced ASCs in the same way as in the \gls{hs} condition. Given the lower number of reports in the \gls{dl}, it is possible that the underlying experiences were qualitatively different, with fewer deep states or psychedelic-like characteristics, and more data would be necessary to make a more robust claim about \gls{dl}'s phenomenology.

Moreover, although not explicitly identified as distinct topics, an examination of individual sentence structures revealed that many reports conveyed emotional valence. Participants also described cognitive changes, including elements of self-reflection and autobiographical memory. Additionally, and likely due to the 360-degree music participants were listening to, auditory experiences were frequently emphasised as an important part of the experience, aligning with the established additive effects of music and stroboscopic stimulation \cite{montgomery_flicker_2024}. Interestingly, these aspects of experience bear some resemblance to findings from a previous study that applied quantitative textual analysis to reports from first-time ayahuasca experiences in ceremonies accompanied by music \cite{cruz_quantitative_2023}.  

\subsection{Methodological Innovation and Implications}

Traditional studies of \glspl{asc} and \glspl{vh} have relied on structured questionnaires such as the Five-Dimensional Altered States of Consciousness Scale (5D-\gls{asc}) \cite{dittrich_standardized_1998,dittrich_5d-asc_2010,studerus_psychometric_2010}, the Phenomenology of Consciousness Inventory (PCI) \cite{pekala_phenomenology_1991} or the Mystical Experience Questionnaire (MEQ) \cite{maclean_factor_2012}. While these instruments facilitate standardised analysis, their predefined categories may overlook unexpected aspects of experience. Moreover, questionnaires can be prone to biases introduced by the framing of questions and the researchers' preconceptions and beliefs about the phenomena being studied. In contrast, open-ended reports allow participants to describe their experiences in their own words, preserving the richness and nuance of their subjective experiences, potentially revealing novel insights. However, analysing such open-ended reports presents significant challenges. The unstructured nature of free-text responses complicates systematic analysis, and interpretation remains susceptible to researcher bias. Practically, processing and analysing open-ended reports is significantly more time-intensive compared to structured methods.. 

To address these challenges while retaining the benefits of open report, we used \gls{nlp} models, to bridge the gap between the richness of qualitative data and the analytical power of quantitative methods. By avoiding predefined categories, our approach enables the discovery of a broader range of experiences, including those not typically associated with \gls{sls} in prior research. The data-driven nature of our approach reduces the influence of researcher preconceptions, allowing patterns to emerge from the data itself. Importantly, our approach is intended to complement rather than replace traditional methods, providing a data-driven pathway to systematically analyse  phenomenological dimensions of subjective experience.

To promote reproducibility and facilitate wider adoption of this approach, we provide an open-access detailed documentation of our MOSAIC pipeline \footnote{https://github.com/romybeaute/MOSAIC}, including preprocessing steps, parameter settings, and interpretation guidelines (see section \ref{sec:fdirections}).

\subsection{Limitations}\label{sec:limitations}
The present study has several important limitations. First, Dreamachine's name and marketing likely influenced participants' expectations, which may have shaped their experiences and subsequent reports. While both \gls{hs} and \gls{dl} versions shared contextual elements (e.g., setting, music), differences in sensory stimulation likely affected participants differently: the \gls{hs} condition's stroboscopic effects may have maintained arousal, whereas the \gls{dl} condition's encouraged a deeply relaxing and potentially hypnagogic state. Moreover, the immersive soundscape in both versions, the guided reflection and integration, as well as the collective nature of the Dreamachine sessions (20--30 people in each session) likely contributed significantly to the emotional depth and cognitive engagement reported by participants. This aligns with research on collective rituals and shared experiences, which have been shown to enhance emotional synchrony and feelings of connectedness among participants \cite{carhart-harris_psychedelics_2018-1,kettner_psychedelic_2021,paez_psychosocial_2015}. It is therefore important to recognise that while \gls{sls} may enhance or catalyse different types of stroboscopically-induced phenomenology, other factors such as set, setting, and individual differences likely play crucial roles in shaping reports of \gls{asc} and complex \glspl{vh} \cite{carhart-harris_psychedelics_2018}. 

A second limitation of this study concerns the sample size of Dreamachine's open reports dataset, relative to the total visitor population ($\sim$40,000). This disparity likely results from the data collection methodology, wherein phenomenological reports were collected through an optional free-text question positioned at the end of a comprehensive digital survey that was one of a number of reflection options available to participants, and limited by availability. Prior to this final open reflection question ("In the Dreamachine I experienced..."), participants had already responded to targeted open-ended questions focusing on specific aspects of their experience, including emotions, visual phenomena, auditory experiences, cognitive changes, and bodily sensations. Consequently, participants may have felt they had already thoroughly documented their experience and opted not to provide additional commentary in the final open reflection, which was the focus of our present analysis.  
To minimise prompt-related biases, and capture a more global account of the entire experience, we analysed only the final open-ended reflections, rather than aggregating multiple free-text responses that could have skewed the results. This approach aimed to capture spontaneous reporting of the most salient aspects of participants' experiences without being guided and influenced by specific prompts about particular phenomena. However, it also resulted in a smaller sample size, as only those particularly motivated to share in-depth reflections contributed. This self-selection may have introduced a sampling bias, as participants who provided detailed written responses may differ systematically from those who did not. This limitation should be considered when interpreting the generalisability of our findings to the broader Dreamachine population. 

Within the \gls{dl} analysis specifically, the significant proportion of unassigned responses (31.32\%) combined with small sample sizes in each identified topic cluster (ranging from n=9 to n=24) presents substantial methodological challenges. These low numbers, particularly when compared to the \gls{hs} clusters, limit the generalisability of findings regarding the prevalence and characteristics of specific experiential types during \gls{dl} sessions. The high proportion of unassigned responses may indicate limitations in our \gls{tm} approach, particularly its ability to capture the full spectrum of \gls{dl} experiences containing minimal experiential content or methodological comments. Future studies would benefit from larger sample sizes to enable more reliable topic identification and stable clustering solutions. Additionally, having phenomenology experts review the topic assignments would help validate the experiential categories identified by the \gls{tm} approach.

From a methodological perspective, clustering and \gls{tm} algorithm stability pose significant challenges due to their sensitivity to initialisation parameters and hyperparameter settings. Despite using grid search for systematic parameter optimisation, the selection of evaluation metrics (e.g., coherence scores) remains inherently subjective, introducing potential researcher bias in determining `optimal' configurations. Additionally, assessing the quality of clustering or \gls{tm} results without a clear ground truth is inherently difficult, as coherence scores may not fully reflect thematic accuracy from a phenomenological perspective. 

Text preprocessing choices also potentially influenced analysis outcomes. While sentence-level tokenization enabled detailed analysis, it may have disrupted contextual coherence across multiple sentences. Finally, topic interpretations were potentially affected by biases inherent to \gls{llm}, particularly given their pre-training on generalised linguistic datasets rather than domain-specific phenomenological reports.  Future studies should incorporate expert human reviews to validate topic interpretations and ensure phenomenological accuracy.

\subsection{Future Directions}\label{sec:fdirections}

Our study opens up several promising avenues for future research. While our current analysis reveals distinct aspects of experiences during \gls{sls} - ranging from simple geometric patterns to more complex narratives or higher-level phenomenological states - the underlying neurocognitive mechanisms remain to be fully elucidated. A critical question in this regard is how expectations and prior knowledge shape the nature and intensity of \gls{sls} experiences.  Future studies could systematically investigate this through expectation manipulation paradigms and measurement of trait suggestibility or phenomenological control, allowing researchers to determine which experiential reports are contextually influenced and which remain stable across varying expectations.

To ensure the robustness and generalisability of these findings, replication studies using varied \gls{sls} setups are needed. Future research should explore how different \gls{sls} frequencies, luminance, durations, and accompanying stimuli affect the types of experiences reported. Methodologically, future studies should focus on developing more robust methods for evaluating the stability and quality of \gls{nlp}-derived topics in the context of subjective experience research. This could include the development of specialised coherence metrics tailored for phenomenological data.

In addition to these empirical directions, our \gls{nlp} approach offers a systematic categorisation of stroboscopically-induced phenomenology through an open source \gls{tm} pipeline. This categorisation provides well-defined phenomenological targets that could be used in future neuroimaging studies of stroboscopically-induced phenomenology to investigate whether different experiential categories (e.g., geometric patterns versus spiritual states) correspond to distinct patterns of brain activity or network connectivity. Establishing such mappings would help advance our understanding of the relationship between subjective experiences and their neural correlates. 

More generally, our \gls{tm} pipeline, which combines BERTopic for semantic clustering and Llama for topic interpretation, offers a systematic, reproducible framework for analysing free-text phenomenological data. The workflow encompasses five key stages: (1) data preprocessing and cleaning, (2) semantic embedding generation, (3) topic modelling and clustering, (4) \gls{llm}-assisted topic interpretation, and (5) hierarchical relationship analysis. 

While our approach largely builds on existing text analysis frameworks, it offers a specific implementation tailored for phenomenological research. While requiring careful consideration of parameters and potential modifications for different datasets, the modular nature of our approach allows researchers to adapt each component - from preprocessing to topic interpretation - to their specific research needs. To facilitate adoption, we provide detailed documentation of our implementation, including preprocessing steps, parameter settings, and interpretation guidelines. By leveraging these \gls{nlp} techniques with \gls{llm}-assisted interpretation, researchers can efficiently process large volumes of textual data without sacrificing the nuanced understanding typically associated with qualitative analysis. Beyond research on \gls{sls}, this approach demonstrates the broader potential of computational methods for analysing qualitative data in consciousness studies and other domains where systematic analysis of subjective reports is valuable.

\section*{CONCLUSION}

Our study makes three primary contributions. First, through \gls{nlp} of Dreamachine experience reports, we identified for the first time categories of experience that traditional questionnaire-based studies of stroboscopic stimulation may have overlooked, particularly in relation to complex hallucinations and \glspl{asc}. While the specific role of \gls{sls} versus contextual influences remains to be determined, our findings reveal a richer range of phenomenological experiences than previously documented in laboratory settings. Second, we demonstrate how combining established \gls{nlp} techniques - \gls{tm} and \glspl{llm} - can effectively bridge qualitative and quantitative approaches in consciousness research and beyond. Our implementation shows that these methods can systematically analyse rich, subjective experiences while preserving their phenomenological complexity, offering a plug-and-play method for identifying neurophenomenological explanatory targets. Third, by providing a fully documented, open-source analytical pipeline, we offer researchers concrete tools to implement similar analyses in their own investigations of phenomenological data. This practical contribution lowers the technical barriers for applying systematic text analysis methods to phenomenological data. Our detailed methodology and open-source pipeline\footnote{\url{ https://github.com/romybeaute/MOSAIC}} provide a guideline for researchers to adapt and apply these techniques to diverse phenomenological datasets, both within consciousness research and in other fields.

\hspace{15cm}

\subsection*{Funding}
RB is funded by the Be.AI Leverhulme scholarship. This project has received funding from the European Research Council (ERC) under the European Union's Horizon 2020 research and innovation programme (grant agreement no. 101019254, project CONSCIOUS) to AKS. DJS is supported by a MRC Grant UKRI083. G.D. was supported by the Institute for Data Valorization, Montreal (IVADO; CF00137433) and the Fonds de recherche du Québec (FRQ; 285289). The Dreamachine Programme was produced by Collective Act, supported by the UK Government and originally commissioned as part of UNBOXED: Creativity in the UK.

\subsection*{Acknowledgements}
We are grateful to everyone who helped realise the Dreamachine programme, and to all the participants who (anonymously) contributed their experiential reports.

\subsection{Conflict of interest}
JC is Founder and Director of Collective Act,  which conceived, produced and delivered the Dreamachine programme. AKS, FM and DJS are scientific collaborators on Dreamachine and received funding from the UK Government via Collective Act, in support of their work. This funding was not related to the present study.

\bibliographystyle{IEEEtran}
\bibliography{topmod}

\end{document}